\documentclass[a4paper, 10pt, conference]{ieeeconf}      
\usepackage{FG2023}
\usepackage{amsmath}
\let\labelindent\relax
\usepackage{enumitem}
\usepackage{adjustbox}
\usepackage{threeparttable}
\usepackage{multirow}

\usepackage{graphicx,wrapfig,lipsum}
\usepackage{floatrow}
\usepackage{caption}
\usepackage{subcaption}

\usepackage{xcolor,colortbl}
\definecolor{Gray}{gray}{0.85}
\newcolumntype{f}{>{\columncolor{Gray}}c}

\FGfinalcopy 

\IEEEoverridecommandlockouts                              
\title{\LARGE \bf
AFFDEX 2.0: A Real-Time Facial Expression Analysis Toolkit
}

\author{\parbox{16cm}{\centering
    {\large Mina Bishay, Kenneth Preston, Matthew Strafuss, Graham Page, Jay Turcot and Mohammad Mavadati}\\
    {\normalsize
    Affectiva, a Smart Eye company\\}}
}

\begin{document}


\ifFGfinal
\thispagestyle{empty}
\pagestyle{empty}
\else
\author{Mina Bishay, Kenneth Preston, Matthew Strafuss, Graham Page, Jay Turcot and Mohammad Mavadati \\ Affectiva, a Smart Eye company\\}
\pagestyle{plain}
\fi
\maketitle

\begin{abstract}

In this paper we introduce AFFDEX 2.0 -- a toolkit for analyzing facial expressions in the wild, that is, it is intended for users aiming to; a) estimate the 3D head pose, b) detect facial Action Units (AUs), c) recognize basic emotions and 2 new emotional states (sentimentality and confusion), and d) detect high-level expressive metrics like blink and attention. AFFDEX 2.0 models are mainly based on Deep Learning, and are trained using a large-scale naturalistic dataset consisting of thousands of participants from different demographic groups. AFFDEX 2.0 is an enhanced version of our previous toolkit \cite{mcduff2016affdex}, that is capable of tracking faces at challenging conditions, detecting more accurately facial expressions, and recognizing new emotional states (sentimentality and confusion). AFFDEX 2.0 outperforms the state-of-the-art methods in AU detection and emotion recognition. AFFDEX 2.0 can process multiple faces in real time, and is working across the Windows and Linux platforms.

\end{abstract}


\section{Introduction}

Facial expression is one of the most informative behaviour in non-verbal communication, that can reveal human affect, emotions, and personality \cite{bartlett2003real}. There are typically two main approaches commonly used for studying facial expressions; the first includes detecting the facial muscle movements (i.e. AUs), described by the Facial Action Coding System (FACS) \cite{EkmanBook97}, while the second interprets the message delivered by a facial expression, where the message is an emotional state (e.g. anger, Joy). Automatic Facial Expression Analysis (AFEA) focuses on detecting expressions described by both approaches, and has been an active research area in Computer Vision in the last 20 years \cite{martinez2017automatic, li2020deep}. AFEA is a vital processing step in a wide range of applications such as ad testing \cite{mcduff2014predicting, mcduff2016applications, mcduff2017new, efremova2020understanding}, driver state monitoring \cite{dua2019autorate, wilhelm2019towards, mualuaescu2019improving}, and health care \cite{girard2014nonverbal, jaiswal2017automatic, bishay2019schinet, bishay2019can}.

Several architectures and toolkits have been developed for AFEA \cite{martinez2017automatic, li2020deep}, however, the datasets that have been used for training most of those architectures have several limitations. First, many datasets were captured in controlled recording conditions (limited illumination levels and camera poses), which subsequently affects the robustness of the AFEA architectures in naturalistic conditions. Second, the number of subjects available in those datasets is relatively limited, making the machine learning models vulnerable to training problems like overfitting. Third, most of the available datasets has participants from specific demographic groups, limiting the AFEA performance on other demographic groups that have not been included in the training. Furthermore, many of those architectures extracted hand-crafted features for AFEA, which have achieved lower performance compared to the deep learning features.  AFEA is also lacking for a reliable and standalone toolkit that is capable of doing the different AFEA tasks (e.g. face detection, head pose estimation, AU detection, emotion recognition, etc) in real time.

\begin{figure*}
  \centering{\includegraphics[width=0.93 \linewidth]{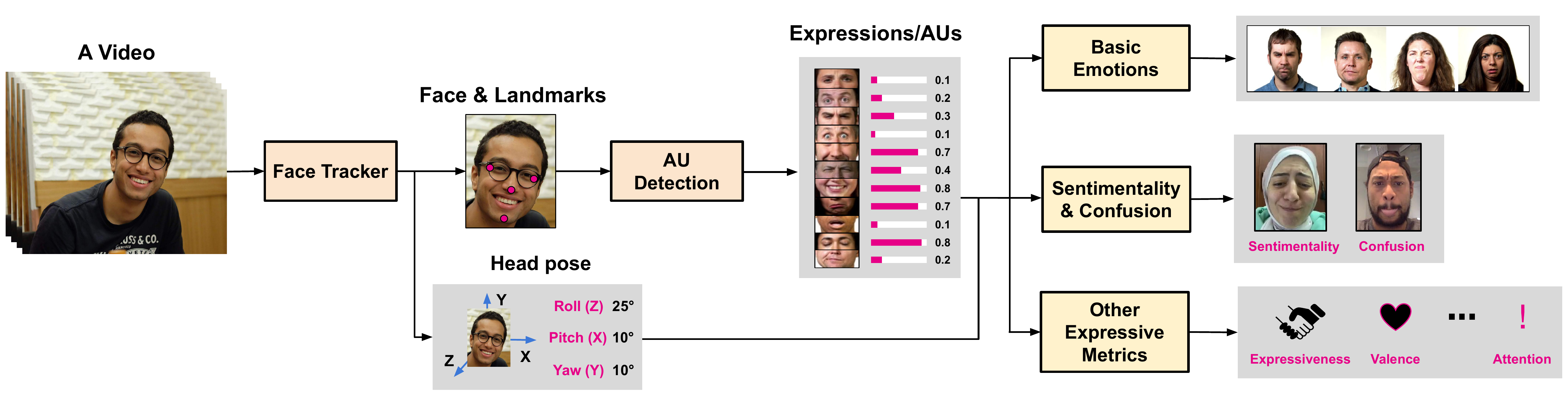}}
  \caption{The full pipeline of AFFDEX 2.0.}
  \label{affdex_sdk_2}
\end{figure*}

In this paper, we build a toolkit (named AFFDEX 2.0) for analyzing facial expressions in the wild. AFFDEX 2.0 can accomplish many of aforementioned AFEA tasks, while overcoming many of the issues highlighted above. Specifically, we use for training and testing our models a large-scale dataset consisting of thousands of videos, that were captured at different recording conditions (i.e. in the wild), and with fully spontaneous facial expressions -- participants in those videos span different demographic groups. Having access to a large-scale dataset allows us to train efficiently Deep Learning based models for facial expression analysis, those models have shown significant improvement compared to AFFDEX 1.0 \cite{mcduff2016affdex}, and other state-of-the-art methods in AFEA. Figure \ref{affdex_sdk_2} shows an overview of AFFDEX 2.0.

%

\begin{table*}
\caption{The face tracking and AU detection average performance across different demographic groups.}
\label{table_tracker_2}
\begin{adjustbox}{width=0.99 \textwidth, center}
  \centering
    \begin{tabular}{|c|c|c|c|c||c|c|c|c|c||c|c|c|c|c||c|c|c|c|c|}
 \hline

\multicolumn{5}{|c|}{\textbf{Age band}} & \multicolumn{5}{|c|}{\textbf{Ethnicity}} & \multicolumn{5}{|c|}{\textbf{Gender}} & \multicolumn{5}{|c|}{\textbf{Glasses}}    \\ \hline  \hline

 & \multicolumn{2}{|c|}{\textbf{Face tracking}} & \multicolumn{2}{|c|}{\textbf{AU detection}} & & \multicolumn{2}{|c|}{\textbf{Face tracking}} & \multicolumn{2}{|c|}{\textbf{AU detection}} &  & \multicolumn{2}{|c|}{\textbf{Face tracking}} & \multicolumn{2}{|c|}{\textbf{AU detection}}  &  &  \multicolumn{2}{|c|}{\textbf{Face tracking}} & \multicolumn{2}{|c|}{\textbf{AU detection}}  \\ \hline \hline 
 
\textbf{AFFDEX} & \textbf{1.0} & \textbf{2.0} & \textbf{1.0} & \textbf{2.0} & \textbf{AFFDEX} & \textbf{1.0} & \textbf{2.0} & \textbf{1.0} & \textbf{2.0} & \textbf{AFFDEX} & \textbf{1.0} & \textbf{2.0} & \textbf{1.0} & \textbf{2.0} & \textbf{AFFDEX} & \textbf{1.0} & \textbf{2.0} & \textbf{1.0} & \textbf{2.0} \\ \hline  \hline 

\textbf{0-17} & 64.4 & 69.3 & 77.7 & 84.1 & \textbf{African} & 85.6 & 94.2 & 76.6 & 86.0 & \textbf{Female} & 93.2 & 96.3 & 83.3 & 87.7 & \textbf{False} & 90.8 & 93.9 & 80.7 & 84.8 \\ \hline 
\textbf{18-24} & 94.9 & 97.7 & 88.4 & 88.7 & \textbf{Caucasian} & 92.4 & 96.1 & 83.2 & 86.6 & \textbf{Male} & 89.7 & 93.2 & 83.3 & 87.0 & \textbf{True} & 91.1 & 95.7 & 86.4 & 88.9 \\ \hline 
\textbf{25-34} & 93.5 & 96.5 & 82.9 & 87.3 & \textbf{East Asian} & 94.0 & 96.4 & 83.5 & 86.6 & \multicolumn{1}{c}{} & \multicolumn{1}{c}{} & \multicolumn{1}{c}{} & \multicolumn{1}{c}{} & \multicolumn{1}{c}{} & \multicolumn{1}{c}{} & \multicolumn{1}{c}{} & \multicolumn{1}{c}{} & \multicolumn{1}{c}{} & \multicolumn{1}{c}{} \\ \cline{1-10} 
\textbf{35-44} & 94.4 & 97.4 & 83.5 & 86.1 & \textbf{Latin} & 96.8 & 98.7 & 83.4 & 85.4 & \multicolumn{1}{c}{} & \multicolumn{1}{c}{} & \multicolumn{1}{c}{} & \multicolumn{1}{c}{} & \multicolumn{1}{c}{} & \multicolumn{1}{c}{} & \multicolumn{1}{c}{} & \multicolumn{1}{c}{} & \multicolumn{1}{c}{} & \multicolumn{1}{c}{}  \\ \cline{1-10} 
\textbf{45-54} & 91.8 & 96.2 & 81.3 & 85.4 & \textbf{South Asian} & 93.8 & 98.0 & 76.4 & 84.3 & \multicolumn{1}{c}{} & \multicolumn{1}{c}{} & \multicolumn{1}{c}{} & \multicolumn{1}{c}{} & \multicolumn{1}{c}{} & \multicolumn{1}{c}{} & \multicolumn{1}{c}{} & \multicolumn{1}{c}{} & \multicolumn{1}{c}{} & \multicolumn{1}{c}{} \\ \cline{1-10} 
\textbf{55-64} & 90.5 & 93.9 & 79.1 & 86.6 & \multicolumn{1}{c}{} & \multicolumn{1}{c}{} & \multicolumn{1}{c}{} & \multicolumn{1}{c}{} & \multicolumn{1}{c}{} & \multicolumn{1}{c}{} & \multicolumn{1}{c}{} & \multicolumn{1}{c}{} & \multicolumn{1}{c}{} & \multicolumn{1}{c}{} & \multicolumn{1}{c}{} & \multicolumn{1}{c}{} & \multicolumn{1}{c}{} & \multicolumn{1}{c}{} & \multicolumn{1}{c}{} \\ \cline{1-5}
\textbf{65+} & 88.1 & 92.0 & 65.4 & 71.0 & \multicolumn{1}{c}{} & \multicolumn{1}{c}{} & \multicolumn{1}{c}{} & \multicolumn{1}{c}{} & \multicolumn{1}{c}{} & \multicolumn{1}{c}{} & \multicolumn{1}{c}{} & \multicolumn{1}{c}{} & \multicolumn{1}{c}{} & \multicolumn{1}{c}{} & \multicolumn{1}{c}{} & \multicolumn{1}{c}{} & \multicolumn{1}{c}{} & \multicolumn{1}{c}{} & \multicolumn{1}{c}{} \\ \cline{1-5} 

    \end{tabular}
\end{adjustbox}
\end{table*}

AFFDEX 2.0 has three main components; face tracker, AU detector, and high-level metrics estimator. First, the face tracker has a deep-based model that is trained on a large dataset for detecting and tracking faces at challenging conditions, as well as estimating the 3D head pose. The AFFDEX 2.0 tracker has better performance (by $\sim$3\%) than our previous toolkit \cite{mcduff2016affdex}. Similarly, the AU detector has a deep-based model that is trained with thousands of videos for detecting 20 AUs. The AU detector boosts the AU detection performance by $\sim$4\% compared to \cite{mcduff2016affdex}. The improved tracking and AU detection performance can be seen across the different demographic groups. In addition, the AU detector shows better performance than the state-of-the-art methods on the DISFA dataset \cite{mavadati2013disfa}.  


The tracker and AU detector predictions are used for estimating some high-level expressive and data quality metrics. These metrics can be categorised into 4 groups; a) 7 basic emotions, b) two new emotional states (sentimentality and confusion), c) other expressive metrics (e.g. blink, valence, attention), and d) data quality metrics (measuring the quality of the detected face images). The different expressive metrics are estimated by aggregating the AU predictions related to the target metric. Our emotion recognition model outperforms the state-of-the-art methods on the AffWild2 dataset \cite{kollias2022abaw}. The inclusion of two additional emotional states (sentimentality and confusion), beyond those typically in published literature, is based on the application area of our technology, i.e. analyzing participants watching advertising and movies, where these are commonly occurring emotional states.

All the models for face tracking, AU detection, and the high-level metrics estimation have been wrapped together in a single SDK (named AFFDEX 2.0). The SDK can process multiple faces in real-time, and can run on two platforms (Windows and Linux). The 3 main components of the SDK; a) face tracking, b) AU detection, and c) high-level metrics estimation will be explained in detail in the following sections. 


\section{Face Tracking}
\label{face_tracker}

The face tracker mainly localizes and tracks faces in a video, and it has two components; a \textit{face detector} that discovers new faces across the video, and a \textit{landmark detector} that tracks the detected faces in every frame. 


For the \textbf{face detector}, we train a deep model based on faster R-CNN \cite{ren2015faster} using a dataset consisting of $\sim$20K images, that is rich in head poses, occlusions, and face sizes. The images have been annotated for multiple faces.  The dataset is splitted into 12K images for training and 7k for validation. For the \textbf{landmark detector}, we train a CNN consisting of 3 convolutional layers for detecting 4 facial landmarks (i.e. outer eye corners, nose tip and chin) on a given face region. The CNN is trained using a dataset collected in the wild, and consisting of $\sim$80K images. The dataset is splitted into 65K images for training and 15k for validation. The detected landmarks are used for estimating the 3D head pose. Figure \ref{affdex_sdk_2} shows the outputs of the face tracker; a) a bounding box including the face, b) the head pose, and c) 4 landmark locations. 


In AFFDEX 2.0, all the video frames are scaled to a fixed resolution, and passed as input to the face detector. Then, the landmark detector is applied on the detected faces in the frames. As the face detector operates on the whole frame and requires a relatively large amount of computations, while the landmark detector operates only on the face region, and has much lower computations than the face detector. Subsequently, we apply the face detector every 0.5s to find lost/new faces, while the landmark detector is applied every frame on the proposals given by the face detector (in the last timestamp being applied). 

\begin{figure*}
  \centering{\includegraphics[width=0.75 \linewidth]{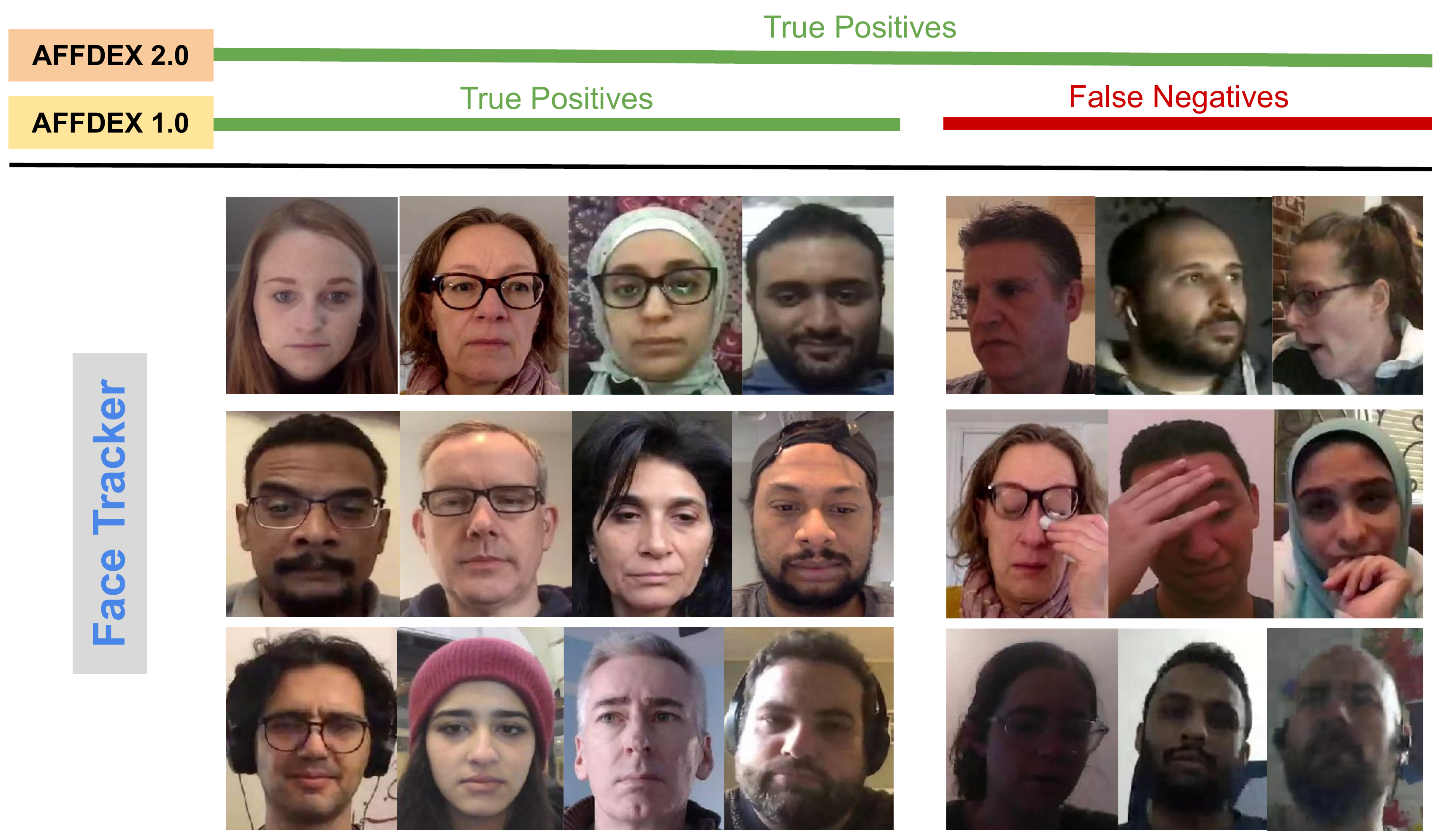}}
  \caption{Qualitative comparison between AFFDEX 2.0 and AFFDEX 1.0 on face tracking. Results show that AFFDEX 2.0 can track faces at more challenging conditions; a) harder head poses (first row), b) hand occlusions (second row), and c) darker illumination (third row).}
  \label{face_tracker_comp}
\end{figure*}

\textbf{Results.} A large in-house testing set consisting of $\sim$8.2K videos ($\sim$6M frames) of participants watching commercial ads worldwide is used for evaluating our face tracker. The testing set has labels for gender, ethnicity, age band, and glasses. The average tracking performance across the testing set has increased from 91.3\% for AFFDEX 1.0 \cite{mcduff2016affdex} to 94.6\%  for AFFDEX 2.0 ($\sim$3\% increase). Table \ref{table_tracker_2} show the tracking performance across the different demographic groups. From Table \ref{table_tracker_2}, we can first conclude that the improvement in face tracking can be seen across all demographic breakdowns. Second, some groups like African ethnicity, people with glasses, and older age bands (55-64, 65+) got largely improved by $\sim$9-12\% in the new tracker. Figure \ref{face_tracker_comp} shows a qualitative comparison between the AFFDEX 1.0 and 2.0 trackers. Figure \ref{face_tracker_comp} highlights how the new tracker can detect faces at more challenging conditions (harder head poses, hand occlusions, darker illumination).

\section{AU detection}

AUs are the building blocks for most of the facial expressions. In this section, we build an AU detector that detects 20 different AUs at naturalistic recording conditions. The AU detector consists of a deep based model that is trained and tested using a large-scale dataset.

\subsection{Dataset.} 

Most of the datasets available in the literature (e.g. DISFA \cite{mavadati2013disfa}, UNBC \cite{lucey2011painful}, BP4D \cite{zhang2014bp4d}) have relatively limited number of participants, recording conditions, and/or diversity in demographics. In our analysis, we use a large-scale dataset that was captured in the wild, and has spontaneous facial expressions. The web-based approach described in \cite{mcduff2013affectiva} is used for collecting thousands of videos for participants watching commercial ads worldwide (from 90+ countries). This dataset has $\sim$55K videos of participants with diverse age, gender and ethnicity. 


The collected videos were annotated for the presence of AUs using trained FACS coders. In addition, videos were labelled for gender, ethnicity, age band, and glasses. For our analysis, we divide the dataset into 40.9K videos for training, 5.9K for validation, and 8.2K for testing. Note that the face tracker and the AU detector have the same testing set. A part of this dataset was made available to the research community through AM-FED \cite{mcduff2013affectiva} and AM-FED+ \cite{mcduff2018fed+}.


\begin{table*}
\caption{The ROC-AUC obtained by the AFFDEX 1.0 \cite{mcduff2016affdex} and AFFDEX 2.0 on the AU testing set.}
\label{table_AUs_1}
\begin{adjustbox}{width=0.99 \textwidth, center}
\begin{threeparttable}
  \centering
\begin{tabular}{|c||c|c|c|c|c|c|c|c|c|c|c|c|c|c|c|c|c|c|c|c||c|}
\hline

\textbf{\ \ \parbox{2.3cm}{ \quad \ Facial \\ Expressions}}  & 
\textbf{\rotatebox[origin=c]{90}{\ \ \parbox{2.3cm}{Inner Brow \\ Raiser (AU1)}}} & \textbf{\rotatebox[origin=c]{90}{\ \ \parbox{2.3cm}{Outer Brow \\ Raiser (AU2)}}} & \textbf{\rotatebox[origin=c]{90}{\ \ \parbox{2.3cm}{Brow Furrow \\ (AU4)}}} & \textbf{\rotatebox[origin=c]{90}{\ \ \parbox{2.3cm}{Upper Lid \\ Raiser (AU5)}}} & \textbf{\rotatebox[origin=c]{90}{\ \ \parbox{2.3cm}{Cheek Raiser \\ (AU6)}}} &
\textbf{\rotatebox[origin=c]{90}{\ \ \parbox{2.3cm}{Lid Tightener \\ (AU7)}}} &
\textbf{\rotatebox[origin=c]{90}{\ \ \parbox{2.3cm}{Nose Wrinkle \\ (AU9)}}} &
\textbf{\rotatebox[origin=c]{90}{\ \ \parbox{2.3cm}{Upper Lip \\ Raiser (AU10)}}} &
\textbf{\rotatebox[origin=c]{90}{\ \ \parbox{2.3cm}{Lip Corner \\ Puller (AU12)}}} &
\textbf{\rotatebox[origin=c]{90}{\ \ \parbox{2.3cm}{Dimpler \\ (AU14)}}} &
\textbf{\rotatebox[origin=c]{90}{\ \ \parbox{2.3cm}{Lip Corner \\ Depressor (AU15)}}} &
\textbf{\rotatebox[origin=c]{90}{\ \ \parbox{2.3cm}{Chin Raiser \\ (AU17)}}} &
\textbf{\rotatebox[origin=c]{90}{\ \ \parbox{2.3cm}{Lip Pucker \\ (AU18)}}} &
\textbf{\rotatebox[origin=c]{90}{\ \ \parbox{2.3cm}{Lip Stretch \\ (AU20)}}} &
\textbf{\rotatebox[origin=c]{90}{\ \ \parbox{2.3cm}{Lip Press \\ (AU24)}}} &
\textbf{\rotatebox[origin=c]{90}{\ \ \parbox{2.3cm}{Lips Part \\ (AU25)}}}   & 
\textbf{\rotatebox[origin=c]{90}{\ \ \parbox{2.3cm}{Jaw Drop \\ (AU26)}}}   & 
\textbf{\rotatebox[origin=c]{90}{\ \ \parbox{2.3cm}{Lip Suck \\ (AU28)}}}   & 
\textbf{\rotatebox[origin=c]{90}{\ \ \parbox{2.3cm}{Eyes Closure \\ (AU43)}}}   & 
\textbf{\rotatebox[origin=c]{90}{\ \ \parbox{2.3cm}{Smirk}}} & 
\textbf{\rotatebox[origin=c]{90}{\ \ \parbox{2.3cm}{Average}}}   \\ \hline \hline

AFFDEX 1.0 \cite{mcduff2016affdex} & 0.76 & 0.79 & 0.86 & \textbf{0.87} & 0.92 & 0.75 & \textbf{0.91} & 0.86 & 0.94 &\textbf{0.86} & 0.78 & 0.79 & 0.91 & 0.86 & 0.76 & 0.86 & 0.63 & 0.91 & \textbf{0.92} &  0.82 & 0.84 \\ \hline

AFFDEX 2.0 & \textbf{0.79} & \textbf{0.84} & \textbf{0.92} & 0.85 & \textbf{0.94} & \textbf{0.83} & 0.89 & \textbf{0.93} & \textbf{0.97} & 0.80 & \textbf{0.88} & \textbf{0.88} & \textbf{0.92} & \textbf{0.93} & \textbf{0.87} & \textbf{0.89} & \textbf{0.71} & \textbf{0.96} & 0.91 & \textbf{0.84} & \textbf{0.88}  \\ \hline

 \end{tabular}
 \begin{tablenotes}
\item Smirk is defined as the asymmetric lip corner puller (AU12) or dimpler (AU14).
\end{tablenotes}
\end{threeparttable}
\end{adjustbox}
\end{table*}

\begin{table*}
\caption{The F1-score obtained by AFFDEX 2.0 and state-of-the-art methods on the DISFA dataset (cross-dataset testing).}
\label{table_disfa_1}
\begin{adjustbox}{width=0.88 \textwidth, center}
  \centering
    \begin{tabular}{|c||c|c|c|c|c|c|c|c|c|}
\hline
 & AFFDEX 2.0 & EB+ \cite{ertugrul2020crossing} & GFT \cite{ertugrul2020crossing} & CDPSL \cite{baltruvsaitis2015cross} & AlexNet \cite{tu2019idennet} & DRML \cite{tu2019idennet} & SVTPT \cite{tu2019idennet} & LightCNN \cite{tu2019idennet} & IdenNet \cite{tu2019idennet}  \\ \hline
AU1 & 0.366 & \textbf{0.571} &  0.370 & - & 0.134 & 0.116 & 0.124 & 0.130 & 0.201  \\ \hline
AU2 & \textbf{0.623} & 0.499 & 0.379 & 0.350 & 0.102 & 0.047 & 0.112 & 0.082 & 0.255  \\ \hline
AU4 & \textbf{0.781} & 0.612 & 0.553 & - & 0.275 & 0.325 & 0.131 & 0.365 & 0.373  \\ \hline
AU6 & \textbf{0.625} & 0.416 & 0.475 & - & 0.337 & 0.328 & 0.259 & 0.413 & 0.496  \\ \hline
AU12 (Smile) & \textbf{0.783} & 0.444 & 0.624 & 0.59 & 0.474 & 0.497 & 0.443 & 0.537 & 0.661  \\ \hline 
AU17 & \textbf{0.832} & - & 0.26 & - & - & - & - & - & - \\ \hline

    \end{tabular}
\end{adjustbox}
\end{table*}
\begin{table*}
\caption{The F1-score obtained by AFFDEX 2.0 and state-of-the-art methods on the DISFA dataset (within-dataset testing).}
\label{table_disfa_2}
\begin{adjustbox}{width=0.88 \textwidth, center}
  \centering
    \begin{tabular}{|c||c|c|c|c|c|c|c|c|c|}
\hline

 & AFFDEX 2.0 & DSIN \cite{corneanu2018deep} & LP \cite{niu2019local} & SRERL \cite{li2019semantic} & EAC \cite{li2018eac} & JAA \cite{shao2018deep} & ARL \cite{shao2019facial} & PT-MT \cite{jacob2021facial} & SEV-NET \cite{yang2021exploiting} \\ \hline
AU1 & 0.366 & 0.424 & 0.299 & 0.457 & 0.415 & 0.437 & 0.439 & 0.461 & 0.553 \\ \hline
AU2 & 0.623 & 0.39 & 0.247 & 0.478 & 0.264 & 0.462 & 0.421 & 0.486 & 0.531 \\ \hline
AU4 & 0.781 & 0.684 & 0.727 & 0.596 & 0.664 & 0.56 & 0.636 & 0.728 & 0.615 \\ \hline
AU6 & 0.625 & 0.286 & 0.468 & 0.471 & 0.507 & 0.414 & 0.418 & 0.567 & 0.536 \\ \hline
AU9 & 0.467 & 0.468 & 0.496 & 0.456 & 0.805 & 0.447 & 0.4 & 0.5 & 0.382 \\ \hline
AU12 (Smile) & 0.783 & 0.708 & 0.729 & 0.735 & 0.893 & 0.696 & 0.762 & 0.721 & 0.716 \\ \hline 
AU25 & 0.832 & 0.904 & 0.938 & 0.843 & 0.889 & 0.883 & 0.952 & 0.908 & 0.957 \\ \hline
AU26 & 0.591 & 0.422 & 0.65 & 0.436 & 0.156 & 0.584 & 0.668 & 0.554 & 0.415 \\ \hline
AVG & \textbf{0.634} & 0.536 & 0.569 & 0.559 & 0.485 & 0.56 & 0.587 & 0.615 & 0.588 \\ \hline

    \end{tabular}
\end{adjustbox}
\end{table*}

\subsection{Modelling} 

Our AU detection pipeline consists of 3 main stages; preprocessing, modeling, and postprocessing. 


In the \textbf{preprocessing} stage, we first detect the face region and the 4 facial landmarks using the face tracker introduced in Section \ref{face_tracker}. Second, the landmarks are used for aligning horizontally the face image. Third, the aligned faces are scaled to a fixed size of 96$\times$96, and passed as an input to two CNNs. 


\textbf{Modeling.} We train two identical CNNs for detecting 20 different AUs. Each CNN has 5 convolutional and 1 fully-connected layers, and is trained for detecting some AUs, using a sampling strategy that compensates for imbalanced AU frequency found in spontaneous data and as a result, our labeled data. In order to avoid the classifier biasing to the most frequent classes, we use two different sampling strategies. The first strategy uses the oversampling technique used in \cite{bishay2017fusing} for sampling an equal number of positive and negative examples across each AU, and subsequently balancing the severely imbalanced AUs. The second strategy samples from our large dataset with a condition of having at least a positive label for one of the AUs in each sampled example. The sampled batches in the second strategy are not totally balanced, however the AU frequency in the training batches has been improved compared to the random sampling, in addition, the sampled batches maintain the correlations incorporated between the different AU labels.

%



Using the first sampling strategy retains the AUs equally represented and balanced in the training batches, but discards the correlations incorporated between the different AUs, while the opposite is the second strategy. Subsequently, we trained two CNNs, one using the first strategy, while the other using the second one. The first strategy is better for AUs with highly imbalanced positive to negative ratio, while the second strategy is better for the rest of the AUs. We use severe augmentation in the training to avoid overfitting problems, and Binary Cross Entropy for calculating the loss. Most of the pre-processing and classification settings are chosen based on the recommendations given in \cite{bishay2021choose, bishay2021cnns}. 


In the \textbf{postprocessing} stage, we apply some operations in order to de-noise the individual frame predictions, as well as compensate for activations caused by individual or environmental differences. Specifically, we first smooth the AU predictions using a 1-dimensional moving mean filter -- this helps in having a more consistent predictions and reducing the noisy AU predictions. Second, the smoothed predictions are normalized by subtracting a baseline classifier output determined by analyzing the previous frames -- this helps in reducing the relatively high activations emerging from objects occluding some participants' faces (e.g glasses, hats, or/and hands). Third, we apply a soft-threshold using a sigmoid function. The purpose of the threshold is two-fold; a) to report a low-false positive rate as our toolkit is required to have reliable predictions, and b) to achieve consistent output between versions of our toolkit. A grid search was conducted to find the best set of postprocessing parameters over the AU validation set. The postprocessing stage typically results in around 2\% improvement in the overall ROC-AUC.




\subsection{Results.} 

\begin{figure*}
  \centering{\includegraphics[width=0.8 \linewidth]{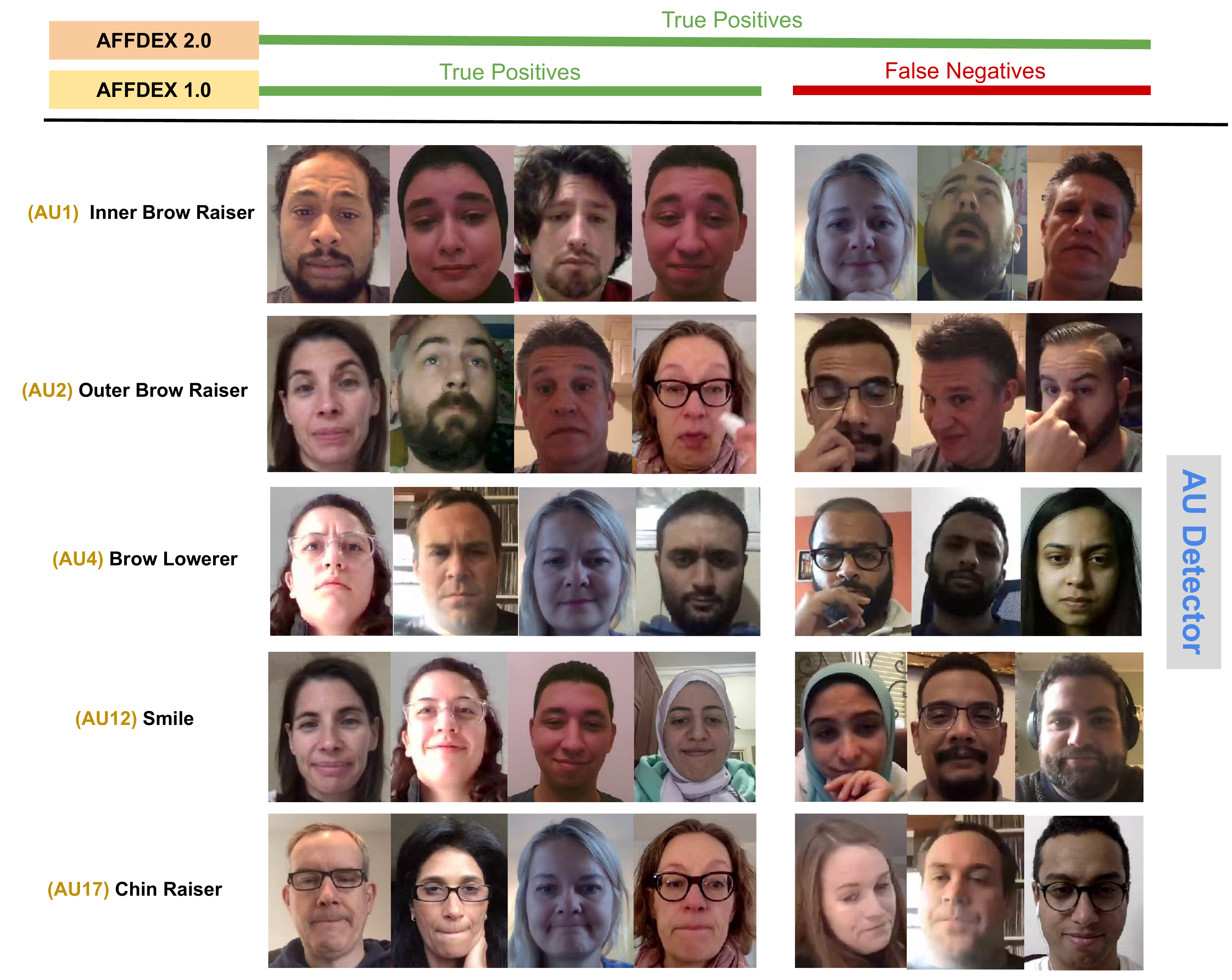}}
  \caption{Qualitative comparison between AFFDEX 2.0 and AFFDEX 1.0 on AU detection.}
  \label{au_detector_comp}
\end{figure*}

We first compare the AU detectors of  AFFDEX 1.0 \cite{mcduff2016affdex} and AFFDEX 2.0 on our large testing set. The area under the ROC curve (ROC-AUC) is used for evaluating the performance. Table \ref{table_AUs_1} shows the performance across the different AUs for both detectors. AFFDEX 2.0 achieves better performance than AFFDEX 1.0 for most of the AUs (by $\sim$4\% on average). Table \ref{table_tracker_2} shows the average performance across the different demographic groups. From Table \ref{table_tracker_2}, we can see that all the groups got improved by AFFDEX 2.0. In addition, some specific groups got largely improved, like the African and South Asian ethnicity who got improved by $\sim$8-9\%. Figure \ref{au_detector_comp} shows a qualitative comparison between the two detectors on 5 AUs (each image has a face with an active AU). AFFDEX 2.0 can detect AUs at more challenging conditions than AFFDEX 1.0. 



Second, we compare our AU detector to the state-of-the-art methods in the literature. For our comparison, we use the DISFA dataset \cite{mavadati2013disfa}, which contains videos recorded for subjects watching short video clips. The DISFA dataset has in total around 130K frames, and was annotated in terms of 12 AUs. Two settings are commonly used for testing, one setting includes testing an architecture on the same dataset used in the training (named \textit{within-dataset testing}), while the other includes testing the architecture on a dataset different from the one used for training (named \textit{cross-dataset testing}). In this paper, we compare our performance to both the methods tested using the within-dataset setting \cite{corneanu2018deep, niu2019local, li2019semantic, li2018eac, shao2018deep, shao2019facial, jacob2021facial, yang2021exploiting}, and those using the cross-dataset setting \cite{ertugrul2020crossing, ertugrul2020crossing, baltruvsaitis2015cross, tu2019idennet}. F1-score is used for evaluating performance, as it is a common KPI reported in the literature. Table \ref{table_disfa_1} and Table \ref{table_disfa_2} compares the performance across the two settings. Note that there are some values missing in Table \ref{table_disfa_1}, as the different state-of-the-art methods have not reported their performance on all the AUs in the DISFA dataset. Although, AFFDEX 2.0 has used a different dataset for training than DISFA, it still can achieve better performance than the state-of-the-art methods in both settings. 


\section{High-Level metrics Estimation}

In this section, we estimate some high-level expressive and data quality metrics based on the tracker and AU detector predictions. These metrics can be categorised into 4 groups; a) basic emotions, b) sentimentality and confusion, c) other expressive metrics, and d) data quality metrics. In the following we will explain how we detect the different high-level metrics.


\subsection{Basic Emotions}

Emotion recognition in AFEA includes mainly detecting 7 basic emotions (Anger, Disgust, Fear, Joy, Sadness, Surprise and Contempt), and the neutral expression. In AFFDEX 2.0, we predict the 7 basic emotions as combinations (i.e. a normalized weighted sum) of the 20 AUs predictions. The used weights are deduced from the Emotional Facial Action Coding System (EMFACS) \cite{friesen1983emfacs}, which defines the combinations of AUs related to each emotion. We add negative weights for opposite AUs so as to minimise the false positives. For example, anger consists of the activation of brow furrow (AU4), lip corner depressor (AU15) and lip pressor (AU24) – in order to reduce the false positives we add negative weights for opposite expressions like inner and outer brow raiser (AU1 and AU2) which are not typically co-firing with brow lowerer (AU4), and the same for smile (AU12) and lip corner depressor (AU15). Finally, the neutral expression is considered active when the 7 basic emotions are absent. 


We evaluate our emotion recognition model on the AffWild2 challenge \cite{kollias2022abaw}. Using the AffWild2 challenge allows all participants to evaluate their methods in similar conditions, for a better and fair comparison. AffWild2 includes around 550 conversational and non-conversational videos, that have been labelled for basic emotions, as well as an \textit{other} emotion category (i.e. representing other non-basic emotions). AFFDEX 2.0 is not tuned for neither the \textit{other} category, nor the conversational emotion analysis, and subsequently we have excluded all the \textit{other} and conversational samples from our analysis. F1-score is used for evaluating performance. Table \ref{table_emotions_1} compares the results obtained by the AFFDEX 2.0 and state-of-the-art methods in the literature \cite{zhang2022transformer, jeong2022facial, xue2022coarse, savchenko2022frame, phan2022expression, kim2022facial, yu2022multi, kollias2022abaw}. Results show that AFFDEX 2.0 outperforms the other methods in emotion recognition, although our model has not been trained on the AffWild2 dataset -- this is in contrast to other methods that have been trained using the AffWild2 dataset.




\subsection{Sentimentality and Confusion}

In AFFDEX 2.0, we introduce and detect two new affective states; sentimentality and confusion, as evoking those emotions is common across different contexts (e.g. watching ads and movies). Sentimentality and confusion models are built on the top of the AU predictions. In the following, we will clarify the dataset and models developed for sentimentality and confusion detection.


\textbf{Datasets.} Affectiva in collaboration with global market agencies have collected thousands of commercial ads. For each ad, several participants were hired to watch the ad, and then fill a survey about how they feel about the ad. A consent was given by the participants to get video recorded while they were watching the ad. Out of the collected ads, experienced ad testers in Affectiva have selected 30 ads (15 sentimental and 15 non-sentimental), and 4.65K participants' videos to form a dataset for sentimentality. Similarly, they have selected 40 ads (19 confusing and 21 non-confusing) and 7.2K participants' videos for confusion analysis. The selected ads for sentimentality and confusion span different markets/countries and have participants with diverse demographics. Non-sentimental and non-confusing ads are typically informative, funny, or musical ads. The sentimental moments in the sentimental ads were labelled by 3 labellers, while for confusion it was hard to find and label specific confusing moments.


\textbf{Sentimentality Detection.} Sentimentality is detected by first analyzing the participants watching the ads using the face tracker and AU detector, and then the activation frequency of the different expressions (i.e. AUs and combinations of AUs) is compared across the sentimental and non-sentimental ads. The expressions are compared using the two ad-level KPIs introduced in \cite{bishay2022automatic}. The first KPI measures how separable is the aggregated sentimentality across sentimental and non-sentimental ads (named ROC-Ad), while the second measures if the aggregated sentimentality is firing high at the right sentimental moments (named ROC-Sent). The two KPIs are calculated on the top of the aggregated sentimentality across different ads. The KPI calculation is based on the area under the ROC curve (ROC-AUC).

\begin{figure}
  \centering{\includegraphics[width=0.95 \linewidth]{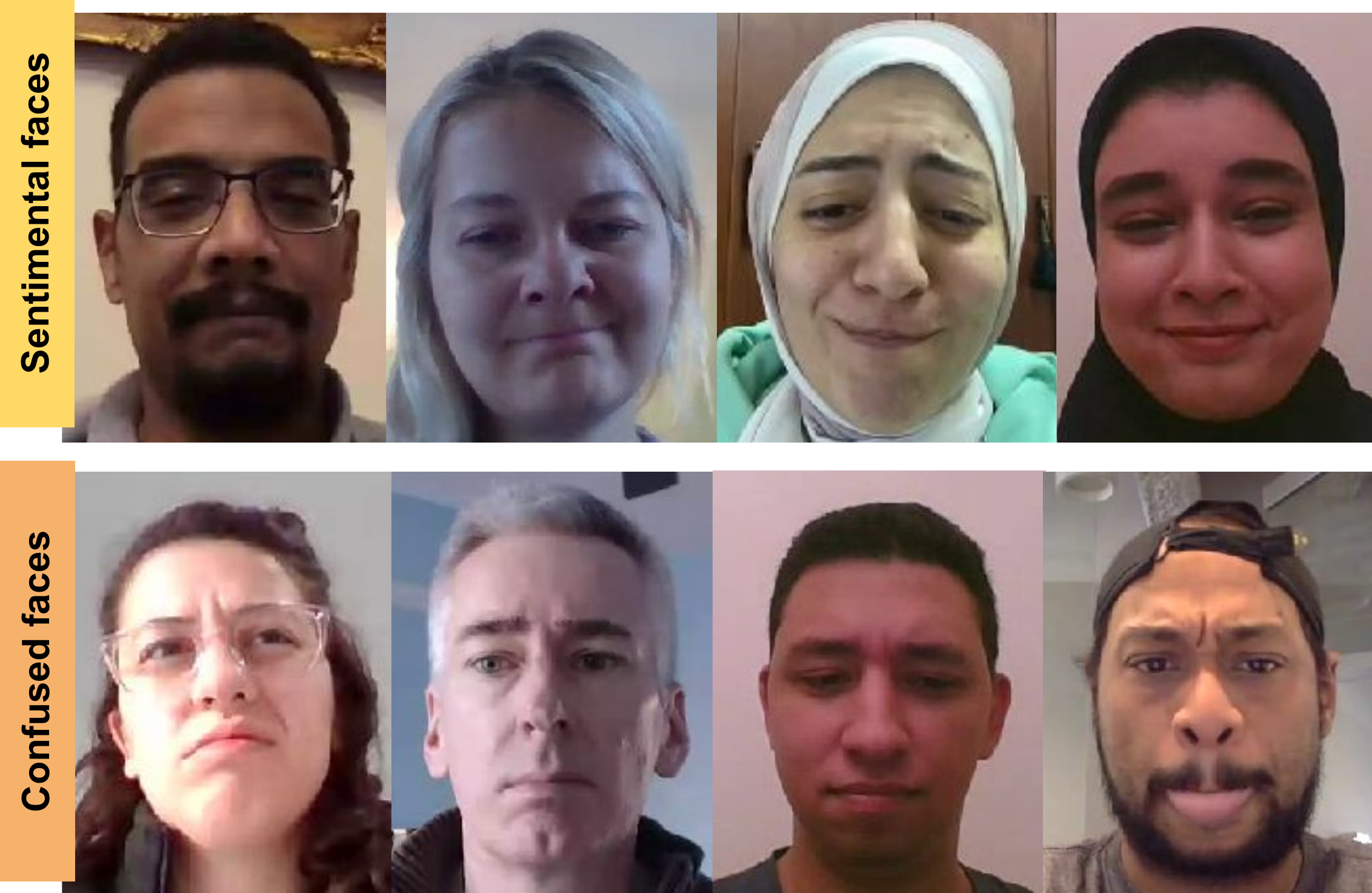}}
  \caption{The positive moments of sentimentality and confusion detected using AFFDEX 2.0.}
  \label{fig_sent_faces}
\end{figure}

By comparing the sentimental and non-sentimental ads, we found 12 combinations of AUs that are significant for sentimentality, those combinations include mostly Joy and sad related AUs. Sentimentality is considered active if any of those 12 combinations were active. We found that combining mainly positive expressions with negative expressions discriminates well sentimental from non-sentimental ads. Some partners used Inner brow raiser (AU1) of AFFDEX 1.0 as a sentimentality score. Table \ref{table_sent_2} compares the performance of our model to the one achieved by using AU1 of AFFDEX 1.0 and 2.0. On average, our sentimentality model has better performance than the chance level and AU1. Figure \ref{fig_sent_faces} (top row) shows some of the detected sentimental faces by AFFDEX 2.0.


\begin{table}
\caption{The F1-score obtained by AFFDEX 2.0 and state-of-the-art methods in emotion recognition on the AffWild2 testing set.}
\label{table_emotions_1}
\begin{adjustbox}{width=0.7 \textwidth, center}
  \centering
    \begin{tabular}{|c||c|}
    \hline

\textbf{Methods} & \textbf{F1 } \\ \hline \hline
\textbf{AFFDEX 2.0} & \textbf{0.360}  \\ \hline
\textbf{Netease Fuxi Virtual Human \cite{zhang2022transformer}} & 0.359 \\ \hline
\textbf{IXLAB \cite{jeong2022facial}}  & 0.338 \\ \hline
\textbf{AlphaAff \cite{xue2022coarse}} & 0.322  \\ \hline

\textbf{HSE-NN \cite{savchenko2022frame}} & 0.303 \\ \hline
\textbf{PRL \cite{phan2022expression}}  & 0.286 \\ \hline
\textbf{dgu \cite{kim2022facial}} & 0.272  \\ \hline
\textbf{USTC-NELSLIP \cite{yu2022multi}} & 0.219 \\ \hline
\textbf{Baseline \cite{kollias2022abaw}}  & 0.205 \\ \hline

    \end{tabular}
\end{adjustbox}
\end{table}

\textbf{Confusion Detection.} For confusion, we repeat the same processing steps used for sentimentality, where we compare the confusing and non-confusing ads in terms of different combinations of AUs, so as to highlight the significant combinations for confusion. We use for the analysis only one KPI (i.e. ROC-Ad), as it was hard to define specific confusing moments in the ads. Comparing the ads shows 6 combinations of AUs that are significant for confusion. In AFFDEX 2.0, the activation of any of those 6 combinations activates confusion. Two of these combinations has brow furrow (AU4), which is similar to other works that have shown a good relation between confusion and AU4 \cite{grafsgaard2011predicting, grafsgaard2011modeling}. Some partners used brow furrow (AU4) of AFFDEX 1.0 as a confusion score. Table \ref{table_sent_2} shows the performance achieved by our confusion model and AU4 detected by AFFDEX 1.0 and 2.0. Our confusion model has better performance than the chance level and AU4. Figure \ref{fig_sent_faces} (bottom row) shows some of the detected confusing faces. Note that all the faces used in the figures belong to Affectiva employees who have been recorded while watching some ads. Both the sentimentality and confusion models shows promising qualitative and quantitative results.

\begin{table}
\caption{Sentimentality and confusion models evaluation.}
\label{table_sent_2}
\begin{adjustbox}{width=0.98 \textwidth, center}
  \centering
    \begin{tabular}{|c||c|c||c|}
 \cline{2-4}

\multicolumn{1}{c}{} & \multicolumn{2}{|c|}{\textbf{Sentimentality}} & \textbf{Confusion} \\ \hline 
\textbf{KPIs} & ROC-Ad & ROC-Sent & ROC-Ad \\ \hline \hline
Random & 50 & 50 & 50 \\ \hline
AFFDEX 1.0 (AU1/AU4) & 71 & 45 & 58 \\ \hline
AFFDEX 2.0 (AU1/AU4) & 71 & 51 & 53 \\ \hline
\textbf{AFFDEX 2.0 (proposed models)} & \textbf{79} & \textbf{60} & \textbf{86}  \\ \hline
    \end{tabular}
\end{adjustbox}
\end{table}



\subsection{Other Expressive Metrics}

In this section we focus on detecting some other expressive metrics. First, the \textit{Blink} is detected when the eye closure (AU43) score gets above and then below a certain threshold during a specific time range. Second, the \textit{Blink Rate} is calculated on the top of the blink output, and it is the number of blinks happening per minute. Third, the participants' \textit{Attention} is estimated by calculating the amount of time the participant spent looking at versus looking away from a screen, specifically turning the head left and right (i.e. yaw angle) is used for calculating attention. Fourth, \textit{Expressiveness} is an overall score of the participant's engagement. The more the face reacts, the more engaged is someone. Expressiveness is calculated as a normalized weighted sum of some upper face AUs (e.g. AU1, AU4) and lower face AUs (e.g. AU12, AU15). Finally, \textit{Valence} is a score for measuring the intrinsic attractiveness or averseness of a situation, and is calculated based on the activation level of the positive expressions (e.g. AU6, AU12) versus the negative expressions (e.g. AU4, AU15).




\subsection{Data Quality Metrics}

AFFDEX 2.0 introduces five metrics for measuring the quality of the detected face images. First, the \textit{Mean Face Luminance} measures how dark or bright is the detected face, and is calculated as the average pixel intensity across the detected face. Second, the \textit{Mean Face Luminance Diff LR} measures the difference in luminance (i.e. mean pixel intensity) between the right and left parts of the face. Third, the \textit{Variance Face Luminance} measures the contrast of the face by calculating the variance in the face pixel intensity. Fourth, the \textit{High Frequency Power} measures the amount of noise available in the face image by summing the square of the high frequency components. Finally, the \textit{Inter Ocular Distance} is the distance between the two eyes, and is calculated by measuring the normalised distance between the 2 landmarks located on the outer eye corners.

\section{Interface}

\begin{figure}
\centering{\includegraphics[width=0.95 \linewidth]{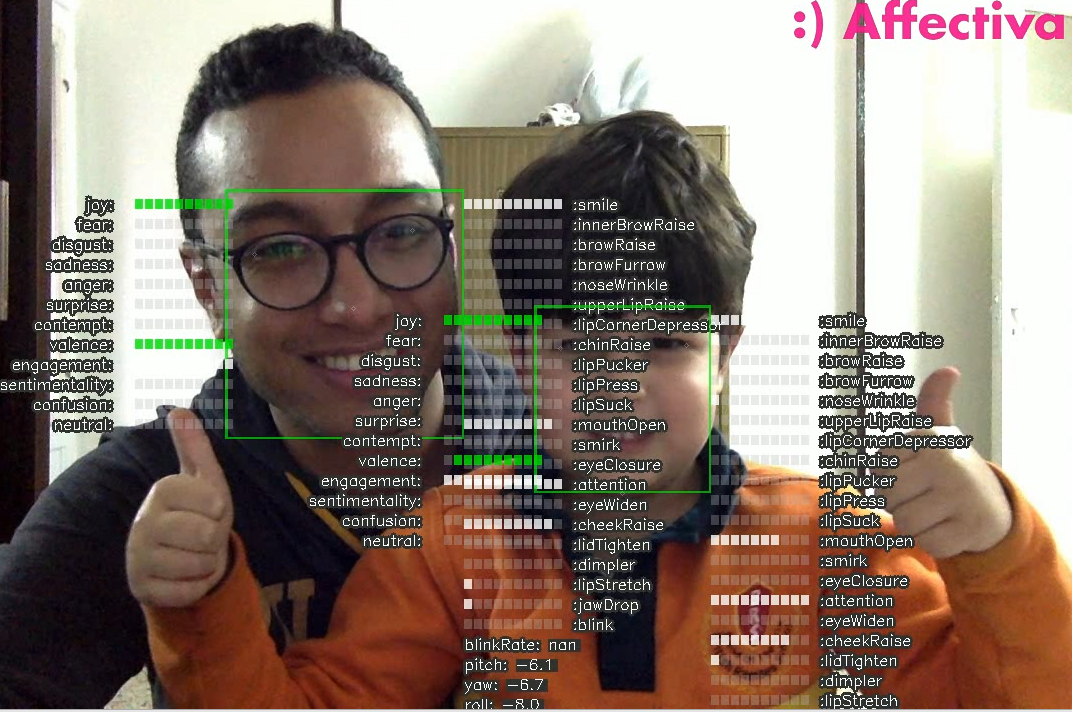}}
\caption{Screenshot of the real-time multi-face expression analysis of AFFDEX 2.0.}
\label{fig_interface}
\end{figure}

AFFDEX 2.0 toolkit is available as an SDK for Windows and Linux platforms. The SDK allows easy integration of the software into other applications. The memory and computational power of a device can impacts the number of faces and frames that can get processed per second. On a Dell-5520 Precision laptop with i7-7820HQ  CPU we can process on average $\sim$62 video Frames per Second (FPS), and $\sim$26-29 FPS using internal and USB webcams with 720p and 1080p resolutions. Figure \ref{fig_interface} shows an example for a desktop demo. The SDK output highlights for each detected face the tracker output (the bounding box and head pose), and the confidence scores for the a) 20 AUs, b) 7 basic emotions, c) sentimentality and confusion, and d) other high-level metrics (e.g. blink and attention), as you can see in Figure \ref{fig_interface}.




\section{Conclusion}

AFFDEX 2.0 is based on a new, more robust face tracker and AU detector, which has deep-based models trained and tested using a large-scale dataset. AFFDEX 2.0 improved the face tracking by $\sim$3\% and the AU detection performance by $\sim$4\%, compared to AFFDEX 1.0. The improved face coverage and AU detection performance is apparent across all demographic groups, and is reflected on the different high-level metrics (e.g. basic emotions, blink, attention). Comparing the performance of AFFDEX 2.0 to other methods in the literature shows that AFFDEX 2.0 achieves the state-of-the-art results in AU detection and emotion recognition. AFFDEX 2.0 also introduces and detects two new emotional states; sentimentality and confusion. Furthermore, AFFDEX 2.0 can process multiple faces in real time, and is working across 2 different platforms (Linux and Windows).


{\small
\bibliographystyle{ieee}
\bibliography{egbib}

\begin{thebibliography}{10}\itemsep=-1pt

\bibitem{baltruvsaitis2015cross}
T.~Baltru{\v{s}}aitis, M.~Mahmoud, and P.~Robinson.
\newblock Cross-dataset learning and person-specific normalisation for
  automatic action unit detection.
\newblock In {\em 2015 11th IEEE International Conference and Workshops on
  Automatic Face and Gesture Recognition (FG)}, volume~6, pages 1--6. IEEE,
  2015.

\bibitem{bartlett2003real}
M.~S. Bartlett et~al.
\newblock Real time face detection and facial expression recognition:
  development and applications to human computer interaction.
\newblock In {\em 2003 Conference on computer vision and pattern recognition
  workshop}, volume~5, pages 53--53. IEEE, 2003.

\bibitem{bishay2021choose}
M.~Bishay, A.~Ghoneim, M.~Ashraf, and M.~Mavadati.
\newblock Choose settings carefully: Comparing action unit detection at
  different settings using a large-scale dataset.
\newblock In {\em 2021 IEEE International Conference on Image Processing
  (ICIP)}, pages 2883--2887. IEEE, 2021.

\bibitem{bishay2021cnns}
M.~Bishay, A.~Ghoneim, M.~Ashraf, and M.~Mavadati.
\newblock Which cnns and training settings to choose for action unit detection?
  a study based on a large-scale dataset.
\newblock {\em arXiv preprint arXiv:2111.08320}, 2021.

\bibitem{bishay2019schinet}
M.~Bishay, P.~Palasek, et~al.
\newblock Schinet: Automatic estimation of symptoms of schizophrenia from
  facial behaviour analysis.
\newblock {\em IEEE Transactions on Affective Computing}, 2019.

\bibitem{bishay2017fusing}
M.~Bishay and I.~Patras.
\newblock Fusing multilabel deep networks for facial action unit detection.
\newblock In {\em Automatic Face \& Gesture Recognition (FG 2017), 2017 12th
  IEEE International Conference on}, pages 681--688. IEEE, 2017.

\bibitem{bishay2019can}
M.~Bishay, S.~Priebe, and I.~Patras.
\newblock Can automatic facial expression analysis be used for treatment
  outcome estimation in schizophrenia?
\newblock In {\em ICASSP 2019-2019 IEEE International Conference on Acoustics,
  Speech and Signal Processing (ICASSP)}, pages 1632--1636. IEEE, 2019.

\bibitem{bishay2022automatic}
M.~Bishay, J.~Turcot, G.~Page, and M.~Mavadati.
\newblock Automatic detection of sentimentality from facial expressions.
\newblock In {\em 2022 IEEE International Conference on Image Processing
  (ICIP)}, pages 321--325. IEEE, 2022.

\bibitem{corneanu2018deep}
C.~Corneanu, M.~Madadi, and S.~Escalera.
\newblock Deep structure inference network for facial action unit recognition.
\newblock In {\em Proceedings of the european conference on computer vision
  (ECCV)}, pages 298--313, 2018.

\bibitem{dua2019autorate}
I.~Dua, A.~U. Nambi, C.~Jawahar, and V.~Padmanabhan.
\newblock Autorate: How attentive is the driver?
\newblock In {\em 2019 14th IEEE International Conference on Automatic Face \&
  Gesture Recognition (FG 2019)}, pages 1--8. IEEE, 2019.

\bibitem{efremova2020understanding}
N.~Efremova, N.~Hajimirza, D.~Bassett, and F.~Thomaz.
\newblock Understanding consumer attention on mobile devices.
\newblock In {\em 2020 15th IEEE International Conference on Automatic Face and
  Gesture Recognition (FG 2020)}, pages 919--919. IEEE, 2020.

\bibitem{EkmanBook97}
P.~Ekman and E.~L. Rosenberg.
\newblock {\em What the face reveals: Basic and applied studies of spontaneous
  expression using the Facial Action Coding System (FACS)}.
\newblock Oxford University Press, USA, 1997.

\bibitem{ertugrul2020crossing}
I.~O. Ertugrul, J.~F. Cohn, L.~A. Jeni, Z.~Zhang, L.~Yin, and Q.~Ji.
\newblock Crossing domains for au coding: Perspectives, approaches, and
  measures.
\newblock {\em IEEE transactions on biometrics, behavior, and identity
  science}, 2(2):158--171, 2020.

\bibitem{friesen1983emfacs}
W.~V. Friesen, P.~Ekman, et~al.
\newblock Emfacs-7: Emotional facial action coding system.
\newblock {\em Unpublished manuscript, University of California at San
  Francisco}, 2(36):1, 1983.

\bibitem{girard2014nonverbal}
J.~M. Girard, J.~F. Cohn, M.~H. Mahoor, S.~M. Mavadati, Z.~Hammal, and D.~P.
  Rosenwald.
\newblock Nonverbal social withdrawal in depression: Evidence from manual and
  automatic analyses.
\newblock {\em Image and vision computing}, 32(10):641--647, 2014.

\bibitem{grafsgaard2011predicting}
J.~F. Grafsgaard, K.~E. Boyer, and J.~C. Lester.
\newblock Predicting facial indicators of confusion with hidden markov models.
\newblock In {\em International Conference on Affective computing and
  intelligent interaction}, pages 97--106. Springer, 2011.

\bibitem{grafsgaard2011modeling}
J.~F. Grafsgaard, K.~E. Boyer, R.~Phillips, and J.~C. Lester.
\newblock Modeling confusion: facial expression, task, and discourse in
  task-oriented tutorial dialogue.
\newblock In {\em International Conference on Artificial Intelligence in
  Education}, pages 98--105. Springer, 2011.

\bibitem{jacob2021facial}
G.~M. Jacob and B.~Stenger.
\newblock Facial action unit detection with transformers.
\newblock In {\em Proceedings of the IEEE/CVF Conference on Computer Vision and
  Pattern Recognition}, pages 7680--7689, 2021.

\bibitem{jaiswal2017automatic}
S.~Jaiswal, M.~F. Valstar, et~al.
\newblock Automatic detection of adhd and asd from expressive behaviour in rgbd
  data.
\newblock In {\em 2017 12th IEEE International Conference on Automatic Face \&
  Gesture Recognition (FG 2017)}, pages 762--769. IEEE, 2017.

\bibitem{jeong2022facial}
J.-Y. Jeong, Y.-G. Hong, D.~Kim, Y.~Jung, and J.-W. Jeong.
\newblock Facial expression recognition based on multi-head cross attention
  network.
\newblock {\em arXiv preprint arXiv:2203.13235}, 2022.

\bibitem{kim2022facial}
J.-H. Kim, N.~Kim, and C.~S. Won.
\newblock Facial expression recognition with swin transformer.
\newblock {\em arXiv preprint arXiv:2203.13472}, 2022.

\bibitem{kollias2022abaw}
D.~Kollias.
\newblock Abaw: Valence-arousal estimation, expression recognition, action unit
  detection \& multi-task learning challenges.
\newblock In {\em Proceedings of the IEEE/CVF Conference on Computer Vision and
  Pattern Recognition}, pages 2328--2336, 2022.

\bibitem{li2019semantic}
G.~Li, X.~Zhu, Y.~Zeng, Q.~Wang, and L.~Lin.
\newblock Semantic relationships guided representation learning for facial
  action unit recognition.
\newblock In {\em Proceedings of the AAAI Conference on Artificial
  Intelligence}, volume~33, pages 8594--8601, 2019.

\bibitem{li2020deep}
S.~Li and W.~Deng.
\newblock Deep facial expression recognition: A survey.
\newblock {\em IEEE Transactions on Affective Computing}, 2020.

\bibitem{li2018eac}
W.~Li, F.~Abtahi, Z.~Zhu, and L.~Yin.
\newblock Eac-net: Deep nets with enhancing and cropping for facial action unit
  detection.
\newblock {\em IEEE transactions on pattern analysis and machine intelligence},
  40(11):2583--2596, 2018.

\bibitem{lucey2011painful}
P.~Lucey, J.~F. Cohn, K.~M. Prkachin, P.~E. Solomon, and I.~Matthews.
\newblock Painful data: The unbc-mcmaster shoulder pain expression archive
  database.
\newblock In {\em Face and Gesture 2011}, pages 57--64. IEEE, 2011.

\bibitem{mualuaescu2019improving}
A.~M{\u{a}}l{\u{a}}escu, L.~C. Du{\c{t}}u, A.~Sultana, D.~Filip, and M.~Ciuc.
\newblock Improving in-car emotion classification by nir database augmentation.
\newblock In {\em 2019 14th IEEE International Conference on Automatic Face \&
  Gesture Recognition (FG 2019)}, pages 1--5. IEEE, 2019.

\bibitem{martinez2017automatic}
B.~Martinez, M.~F. Valstar, et~al.
\newblock Automatic analysis of facial actions: A survey.
\newblock {\em IEEE Transactions on Affective Computing}, 2017.

\bibitem{mavadati2013disfa}
S.~M. Mavadati, M.~H. Mahoor, et~al.
\newblock Disfa: A spontaneous facial action intensity database.
\newblock {\em IEEE Transactions on Affective Computing}, 4(2):151--160, 2013.

\bibitem{mcduff2017new}
D.~McDuff.
\newblock New methods for measuring advertising efficacy.
\newblock In {\em Digital Advertising}, pages 327--342. Routledge, 2017.

\bibitem{mcduff2018fed+}
D.~McDuff, M.~Amr, and R.~El~Kaliouby.
\newblock Am-fed+: An extended dataset of naturalistic facial expressions
  collected in everyday settings.
\newblock {\em IEEE Transactions on Affective Computing}, 10(1):7--17, 2018.

\bibitem{mcduff2016applications}
D.~McDuff and R.~El~Kaliouby.
\newblock Applications of automated facial coding in media measurement.
\newblock {\em IEEE transactions on affective computing}, 8(2):148--160, 2016.

\bibitem{mcduff2014predicting}
D.~McDuff, R.~El~Kaliouby, J.~F. Cohn, and R.~W. Picard.
\newblock Predicting ad liking and purchase intent: Large-scale analysis of
  facial responses to ads.
\newblock {\em IEEE Transactions on Affective Computing}, 6(3):223--235, 2014.

\bibitem{mcduff2013affectiva}
D.~McDuff, R.~Kaliouby, et~al.
\newblock Affectiva-mit facial expression dataset (am-fed): Naturalistic and
  spontaneous facial expressions collected.
\newblock In {\em Proceedings of the IEEE Conference on Computer Vision and
  Pattern Recognition Workshops}, pages 881--888, 2013.

\bibitem{mcduff2016affdex}
D.~McDuff, A.~Mahmoud, M.~Mavadati, et~al.
\newblock Affdex sdk: a cross-platform real-time multi-face expression
  recognition toolkit.
\newblock In {\em Proceedings of the 2016 CHI conference extended abstracts on
  human factors in computing systems}, pages 3723--3726, 2016.

\bibitem{niu2019local}
X.~Niu, H.~Han, S.~Yang, Y.~Huang, and S.~Shan.
\newblock Local relationship learning with person-specific shape regularization
  for facial action unit detection.
\newblock In {\em Proceedings of the IEEE/CVF Conference on computer vision and
  pattern recognition}, pages 11917--11926, 2019.

\bibitem{phan2022expression}
K.~N. Phan, H.-H. Nguyen, V.-T. Huynh, and S.-H. Kim.
\newblock Expression classification using concatenation of deep neural network
  for the 3rd abaw3 competition.
\newblock {\em arXiv preprint arXiv:2203.12899}, 2022.

\bibitem{ren2015faster}
S.~Ren, K.~He, R.~Girshick, and J.~Sun.
\newblock Faster r-cnn: Towards real-time object detection with region proposal
  networks.
\newblock {\em Advances in neural information processing systems}, 28:91--99,
  2015.

\bibitem{savchenko2022frame}
A.~V. Savchenko.
\newblock Frame-level prediction of facial expressions, valence, arousal and
  action units for mobile devices.
\newblock {\em arXiv preprint arXiv:2203.13436}, 2022.

\bibitem{shao2018deep}
Z.~Shao, Z.~Liu, J.~Cai, and L.~Ma.
\newblock Deep adaptive attention for joint facial action unit detection and
  face alignment.
\newblock In {\em Proceedings of the European conference on computer vision
  (ECCV)}, pages 705--720, 2018.

\bibitem{shao2019facial}
Z.~Shao, Z.~Liu, J.~Cai, Y.~Wu, and L.~Ma.
\newblock Facial action unit detection using attention and relation learning.
\newblock {\em IEEE transactions on affective computing}, 2019.

\bibitem{tu2019idennet}
C.-H. Tu, C.-Y. Yang, and J.~Y.-j. Hsu.
\newblock Idennet: Identity-aware facial action unit detection.
\newblock In {\em FG 2019}, pages 1--8. IEEE, 2019.

\bibitem{wilhelm2019towards}
T.~Wilhelm.
\newblock Towards facial expression analysis in a driver assistance system.
\newblock In {\em 2019 14th IEEE International Conference on Automatic Face \&
  Gesture Recognition (FG 2019)}, pages 1--4. IEEE, 2019.

\bibitem{xue2022coarse}
F.~Xue, Z.~Tan, Y.~Zhu, Z.~Ma, and G.~Guo.
\newblock Coarse-to-fine cascaded networks with smooth predicting for video
  facial expression recognition.
\newblock In {\em Proceedings of the IEEE/CVF Conference on Computer Vision and
  Pattern Recognition}, pages 2412--2418, 2022.

\bibitem{yang2021exploiting}
H.~Yang, L.~Yin, Y.~Zhou, and J.~Gu.
\newblock Exploiting semantic embedding and visual feature for facial action
  unit detection.
\newblock In {\em Proceedings of the IEEE/CVF Conference on Computer Vision and
  Pattern Recognition}, pages 10482--10491, 2021.

\bibitem{yu2022multi}
J.~Yu, Z.~Cai, P.~He, G.~Xie, and Q.~Ling.
\newblock Multi-model ensemble learning method for human expression
  recognition.
\newblock {\em arXiv preprint arXiv:2203.14466}, 2022.

\bibitem{zhang2022transformer}
W.~Zhang, F.~Qiu, S.~Wang, H.~Zeng, Z.~Zhang, R.~An, B.~Ma, and Y.~Ding.
\newblock Transformer-based multimodal information fusion for facial expression
  analysis.
\newblock In {\em Proceedings of the IEEE/CVF Conference on Computer Vision and
  Pattern Recognition}, pages 2428--2437, 2022.

\bibitem{zhang2014bp4d}
X.~Zhang, L.~Yin, et~al.
\newblock Bp4d-spontaneous: a high-resolution spontaneous 3d dynamic facial
  expression database.
\newblock {\em Image and Vision Computing}, 32(10):692--706, 2014.

\end{thebibliography}
}

\end{document}